\documentclass[10pt, conference, letterpaper]{IEEEtran} 
\IEEEoverridecommandlockouts
\usepackage{cite}
\usepackage{amsmath,amssymb,amsfonts}
\usepackage{graphicx}
\usepackage{textcomp}
\usepackage{xcolor}
\usepackage{listings}
\usepackage{booktabs}
\usepackage{multirow}
\usepackage{xurl}
\usepackage{hyperref}
\usepackage{algorithm}
\usepackage{nicematrix}
\usepackage{tikz}

\usepackage{algpseudocode}
\usepackage{subcaption}
\usepackage[font=it,labelfont=bf]{caption}

\usepackage{flushend}
\usepackage{capt-of}

\usepackage{pifont}
\newcommand{\cmark}{\textcolor{green!60!black}{\ding{51}}}
\newcommand{\gcmark}{\textcolor{black!40}{\ding{51}}}
\newcommand{\xmark}{\textcolor{red!80!black}{\ding{55}}}

\definecolor{warningred}{RGB}{220, 53, 69}       
\definecolor{warningorange}{RGB}{0, 123, 255}    
\definecolor{warningpurple}{RGB}{138, 43, 226}    
\definecolor{warninggreen}{RGB}{70, 161, 4}       

\newif\ifcomment

\commenttrue 

\ifcomment
\newcommand{\commentmcv}[1]{{\emph{\textcolor{warningred}{[MCV: #1]}}}}
\newcommand{\commental}[1]{{\emph{\textcolor{warningorange}{[Alan: #1]}}}}
\newcommand{\commentab}[1]{{\emph{\textcolor{warningpurple}{[Avhi: #1]}}}}
\newcommand{\commentap}[1]{{\emph{\textcolor{warninggreen}{[Apala: #1]}}}}
\newcommand{\commentmik}[1]{{\emph{\textcolor{brown}{[Mikhail: #1]}}}}

\else
\newcommand{\commentmcv}[1]{}
\newcommand{\commental}[1]{}
\newcommand{\commentab}[1]{}
\newcommand{\commentap}[1]{}
\newcommand{\commentmik}[1]{}
\fi

\begin{document}

\title{SLED: Sensor SLicing from Edge to Datacenter}
\author{
\IEEEauthorblockN{
Alan Subedi$^{*,1}$\quad
Avhishek Biswas$^{*,1}$\quad
Apala Pramanik$^{1}$\quad
Amlan Balabantaray$^{2}$\quad
Mikhail Titov$^{3}$\quad \\
Santosh Pitla$^{2}$\quad
Ozgur Kilic$^{3}$\quad
Mehmet C. Vuran$^{1}$
}
\IEEEauthorblockA{$^1$\textit{School of Computing,
University of Nebraska--Lincoln}, Lincoln, NE, USA\\
\{asubedi8, abiswas3, apramanik2\}@huskers.unl.edu,
mcv@unl.edu}
\IEEEauthorblockA{$^2$\textit{Department of Biological Systems Engineering,
University of Nebraska--Lincoln}, Lincoln, NE 68583, USA\\
abalabantaray2@huskers.unl.edu, spitla@unl.edu}
\IEEEauthorblockA{$^3$\textit{Brookhaven National Laboratory},
Upton, NY 11973, USA\\
\{mtitov, okilic\}@bnl.gov}
\thanks{$^*$Alan Subedi and Avhishek Biswas contributed equally to this work.}
}

\maketitle
\bstctlcite{IEEEexample:BSTcontrol}

\begin{abstract}

Modern sensing systems generate large volumes of heterogeneous data streams from multiple sensor modalities, increasing reliance on datacenters for computation, artificial intelligence/superintelligence (AI/SI) inference, and AI/SI training. Each sensor stream may have different network and compute requirements depending on its type and data volume. However, current wireless systems typically allocate resources at the device level, leading to contention among co-located sensor streams, where bulk data transfers may starve latency-critical telemetry. To this end, this paper introduces Sensor Slicing (SL). SL provides deterministic, medium access control (MAC)-level isolation for individual sensors within a single user equipment (UE) in open radio access networks (O-RAN). Using agentic coding techniques, SL is implemented in two O-RAN-compliant packages, reducing development time by $400$ times (2\,h\,40\,min [OCUDU] vs.\ 6.5~weeks [OAI]), 228-fold fewer human prompts (4 vs.\ 913), and less than half the tokens (0.64B vs.\ 1.43B). An associated end-to-end framework, \textit{SLED}, is implemented to establish per-sensor radio and network isolation across edge, campus, and remote high-performance computing (HPC) tiers. Over-the-air wireless experiments demonstrate that while conventional network slicing causes critical delays to increase more than $2,300$-fold under high traffic load, \textit{SLED} maintains stable latency. Scaling evaluations on the two RAN stacks show that throughput remains within $0.9\%$ and $1.4\%$ of expected allocations even as the number of sensor slices increases. The results demonstrate that \textit{SLED} preserves latency-critical telemetry while effectively utilizing resources across the sensor-to-HPC pipeline. The \textit{SLED} software and evaluation datasets are publicly available at \href{https://github.com/UNL-CPN-Lab/SLED-Sensor-SLicing-from-Edge-to-Datacenter}{https://github.com/UNL-CPN-Lab/SLED}

\end{abstract}
\begin{IEEEkeywords}
Sensor slicing, network slicing, 6G RAN, Open RAN, sensor-to-HPC pipeline, real-time inference.

\end{IEEEkeywords}


\section{Introduction}
\label{sec:introduction}
With the proliferation of artificial intelligence/super intelligence (AI/SI) approaches, modern cyber-physical systems increasingly offload sensing workloads from resource-constrained field devices to edge computing, cloud computing, and high performance computing (HPC) resources~\cite{mach2017mobile,kurafeeva2025xgfabric}. A single platform (e.g., field robot, vehicle, or scientific equipment) can generate multiple sensor modalities (e.g., image, telemetry, bulk data) with different rate, latency, reliability, and privacy requirements. As these systems scale to fleets, the sensor-to-HPC traffic becomes more heterogeneous and numerous~\cite{albiero2022swarm}. Accordingly, wireless interconnect between field devices and HPC becomes a key system constraint, where sharing a common wireless resource can introduce contention that degrades HPC throughput and affects end-to-end round-trip times.


Commercial cellular or Wi-Fi networks operating on shared or unlicensed spectrum provide limited isolation between heterogeneous traffic, allowing sensing data to contend with latency-critical traffic. Private cellular networks offer a software-defined control plane that can isolate competing traffic flows through \textit{network slicing}, and guarantee network slice-specific quality of service~\cite{popovski20185g,3gpp2026}. This is particularly challenging in the uplink (from sensor to base station), where the resources granted by the base station are arbitrated by the user equipment (UE) through logical channel prioritization (LCP)~\cite{3gpp2026}. However, LCP does not enforce weighted proportional capacity guarantees across multiple streams. Existing approaches provide per-network-slice control~\cite{cheng2024oranslice}, but do not provide end-to-end, per-sensor isolation across both the uplink (UL) and the downlink (DL). Accordingly, latency-critical telemetry can be starved by bulk transfers, directly impacting an HPC inference workload performance. This creates a critical benchmarking and performance modeling gap, where existing network slicing mechanisms do not provide deterministic, per-sensor resource guarantees at the user equipment (UE) level to guarantee HPC application performance. 
\begin{figure}[t]
\centering
\includegraphics[width=0.95\columnwidth]{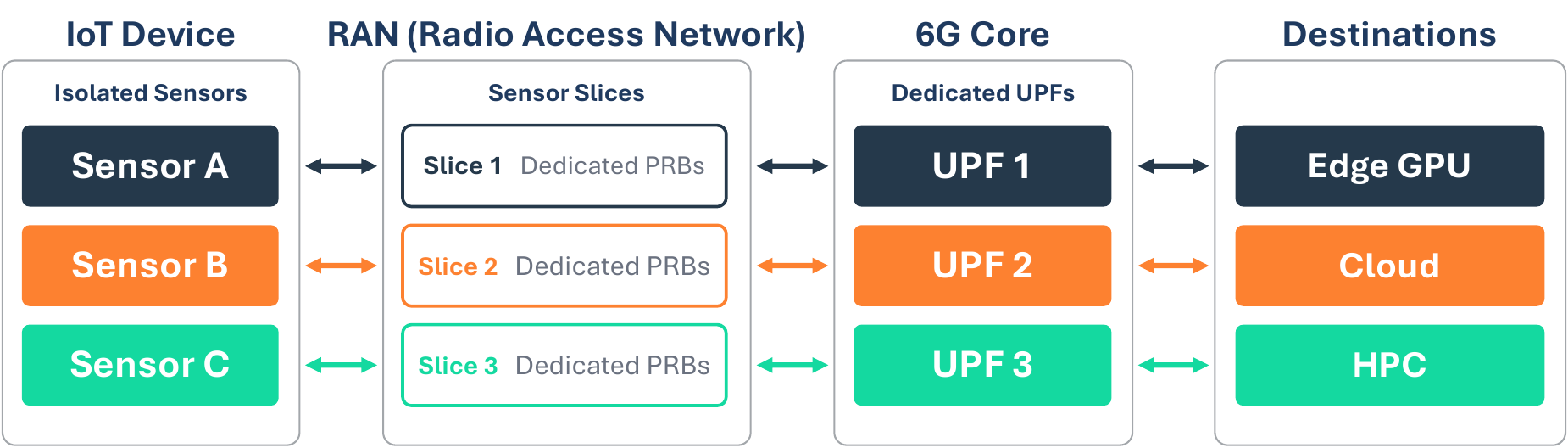}
\caption{The field-to-HPC pipeline. A sensor-sliced private network couples heterogeneous sensor streams with distributed compute tiers.}
\vspace{-0.2in}
\label{fig:overview}
\end{figure}

We present \textit{SLED} (\textbf{S}ensor S\textbf{L}icing from \textbf{E}dge to \textbf{D}atacenter), an end-to-end framework designed to allocate wireless, network, and compute resources between edge, cloud, and HPC resources as shown in Fig.~\ref{fig:overview}. \textit{SLED} virtualizes a physical sensing platform into multiple logical sensors, each with its own sensor slice identity and radio budget. We introduce \textit{Sensor Slicing}, which is a fine-grained radio access network (RAN) scheduling mechanism that uses sensor slice requirements to enforce per-sensor radio resource allocation on a shared UE. Accordingly, sensor streams are isolated at the medium access control (MAC) level and coupled end-to-end with their compute placement. We implement the scheduler through an agentic coding workflow on two open-source O-RAN-compliant packages,  using simulation-based validation to verify each implementation before hardware deployment. We validate \textit{SLED} in real-world with an over-the-air (OTA) testbed.

Our contributions are as follows:
\begin{enumerate}
    \item \textbf{Sensor Slicing, MAC-level intra-UE per-sensor isolation.} We introduce \textit{Sensor Slicing} to assign each sensor stream its own slice identity and radio budget. To this end, two scheduling algorithms are developed: Credit-based slot allocation at the downlink, and deterministic slot ownership at the uplink.
    \item \textbf{\textit{SLED}, an end-to-end sensor-to-compute pipeline.} \textit{SLED} couples multi-modal field sensor streams over a sensor-sliced private RAN with diverse edge, cloud, HPC, and leadership computing facility (LCF) resources for real-time inference.
    \item \textbf{An agentic-AI/SI build-and-verify workflow for RAN development.} The sensor slicing scheduler is implemented by a coding agent in Open-RAN-compliant packages, OpenAirInterface (OAI)~\cite{oai} and OCUDU~\cite{ocudu} with simulation-based gates and automated acceptance tests before hardware deployment, while a human orchestrator controls the design and OTA testing.
    \item \textbf{Open-source Software:} The \textit{SLED} software and evaluation datasets are publicly available\footnote{\href{https://github.com/UNL-CPN-Lab/SLED-Sensor-SLicing-from-Edge-to-Datacenter}{https://github.com/UNL-CPN-Lab/SLED}}.
\end{enumerate}

In the remainder of the paper, we review related work in Sec. ~\ref{sec:related_work} and describe the end-to-end pipeline in Sec.~\ref{sec:pipeline}, followed by the per-sensor slicing mechanism and the implementation in Sec.~\ref{sec:implementation}. In Sec.~\ref{sec:evaluation}, SLED evaluations are presented, and the paper is concluded in Sec.~\ref{sec:conclusion}.

\section{Related Work}
\label{sec:related_work}
We review related work in three areas: RAN Slicing, Agentic RAN Development, and End-to-End Sensor-to-HPC pipeline.

\noindent \textbf{RAN Slicing:} In 5G systems, the research on RAN slicing primarily focuses on radio resource optimization~\cite{polese2023coloran,dorcheh2025dora, barker2025real,tsampazi2024pandora}. In~\cite{zhao2025adaslicing}, online learning adapts radio resource allocations to changing network conditions, improving resource utilization while meeting each slice's performance requirements. Reinforcement learning methods in~\cite{intentaware2024} allocate radio resources to slices based on their service requirements. From a system and evaluation perspective, these approaches generally rely on existing network slicing capabilities in current open-source RAN stacks, such as OpenAirInterface (OAI)~\cite{oai}, srsRAN~\cite{srsran}, and OCUDU~\cite{ocudu}. RAN key performance indicators (KPIs) and machine learning (ML) are used, often through xApps, running on the near-RT RIC. In these stacks, currently, slicing is implemented at the UE level, with one network slice assigned to each UE. Accordingly, all workloads on a UE use the same slice and share the allocated radio resources.

A few approaches focus on expanding slicing capabilities at a system level. An early effort in~\cite{foukas2017orion} introduces slicing capabilities into open-source RAN systems by extending OAI's LTE base station to support per-UE network slicing. The MAC scheduler is extended in FlexSlice~\cite{chen2023flexslice} to support intra-slice scheduling for downlink transmissions. However, these implementations limit one network slice to serve a UE. 

In fact, the 5G standard allows for a single UE to be served by multiple network slice instances simultaneously~\cite{3gpp_23501}. This is governed through Network Slice Selection Assistance Information (NSSAI) and the allowed NSSAI set informed by the gNB to the UE. However, the standard does not ensure deterministic isolation at the MAC-level. Moreover, we are not aware of a full open-source implementation of these features. The work closest to ours is ORANSlice~\cite{cheng2024oranslice}, where the OAI MAC scheduler is modified to support up to two slices on the same UE in the downlink. However, ORANSlice does not support uplink intra-UE slicing, preventing isolation for the majority of network traffic (e.g., TCP, HTTP), and supports only two slices per UE. Furthermore, at the MAC schedule level, only one slice is allocated per slot, where the Physical Resource Blocks (PRBs), the fundamental radio resource block in 5G NR, in that slot are allocated based on a pre-determined slice ratio. This leaves the remaining PRBs in that slot unused, increasing latency and reducing resource efficiency, as shown in Sec.~\ref{sec:evaluation}. 

\begin{table}[t!]
\centering
\caption{Comparison of network and sensor slicing capabilities and supported traffic in State-of-the-Art vs.~Sensor Slicing.}
\label{tab:capability}
\scriptsize
\setlength{\tabcolsep}{2pt}
\renewcommand{\arraystretch}{1.05}
\resizebox{0.98\columnwidth}{!}{%
\begin{tabular}{@{}lccccc|c@{}}
\toprule
\textbf{Capability} &
\shortstack{\textbf{srsRAN}\\\cite{srsran}} &
\shortstack{\textbf{OCUDU}\\\cite{ocudu}} &
\shortstack{\textbf{OAI}\\\cite{oai}} &
\shortstack{\textbf{ORANSlice}\\\cite{cheng2024oranslice}} &
\shortstack{\textbf{Sensor}\\\textbf{Slicing}} &
\shortstack{\textbf{5G Std.}\\\cite{3gpp_23501}} \\
\midrule
\textbf{\textit{Network Slicing}}
  & \cmark & \cmark & \cmark & \cmark & \cmark & \gcmark \\
\midrule
\multicolumn{7}{@{}l}{\textbf{\textit{Sensor Slicing (SL)}}} \\
\quad Downlink
  & \xmark & \xmark & \xmark & \cmark & \cmark & \gcmark \\
\quad Uplink
  & \xmark & \xmark & \xmark & \xmark & \cmark & \gcmark \\
\quad \shortstack[l]{Deterministic\\isolation}
  & \xmark & \xmark & \xmark & \xmark & \cmark & \xmark \\
\quad Max. \# slices
  & -- & -- & -- & 2 & 7 & \textcolor{black!40}{8} \\
\midrule
\multicolumn{7}{@{}l}{\textbf{\textit{SL traffic support}}} \\
\quad UDP Downlink
  & -- & -- & -- & \cmark & \cmark & \gcmark \\
\quad UDP Uplink
  & -- & -- & -- & \xmark & \cmark & \gcmark \\
\quad TCP
  & -- & -- & -- & \xmark & \cmark & \gcmark \\
\bottomrule
\end{tabular}%
}
\vspace{-0.2in}
\end{table}


With \textit{SLED}, we bridge this gap between 5G multi-slice support and practical end-to-end implementations (Table~\ref{tab:capability}) by introducing Sensor Slicing that includes two novel scheduling mechanisms enabling isolation both at the downlink and the \textit{uplink}, and supporting a large number of sensor slices within a single UE. 
\noindent \textbf{Agentic RAN Development:} Recently, Large Language Models (LLMs) have been employed to support RAN development, testing, and operation. In~\cite{wu2025llmxapp}, LLMs are used for radio resource optimization, while an intrusion detection system is built with LLMs in~\cite{moore2025integrated}. Moreover, the protocol validation process is automated in~\cite{ai5gtest} by capturing signaling messages and comparing them with the 3GPP specifications. Furthermore, an open weight model is fine tuned in~\cite{oransight} to improve ORAN-specific knowledge. LLMs are utilized for RAN code modification in~\cite{elkael2026allstar} to implement existing scheduling algorithms from research papers and perform OTA validation. However, in these studies, LLMs are used for a specific task within an existing RAN, such as resource allocation, testing, or implementing a known scheduling algorithm. In contrast, we use an agentic workflow to develop a new RAN capability over an extended development period, where the agent iteratively implements, builds, and tests each version.


\noindent \textbf{End-to-End Sensor to HPC Pipeline:} There are very few studies on end-to-end sensor to HPC pipelines with RAN slicing. The closest is xGFabric~\cite{kurafeeva2025xgfabric}, where private 5G network slicing is employed to connect field sensors to HPC facilities. However, in the xGFabric implementation, isolation is limited at the UE level.
\section{End-to-End Sensor-to-HPC Pipeline}
\label{sec:pipeline}
In this section, we present the \textit{SLED} pipeline and its end-to-end slice enforcement. We first describe the private 5G/6G RAN architecture, then explain sensor-level slicing and its DL/UL enforcement, and finally describe how each slice is carried from the core to its assigned compute tier.
\subsection{Private 5G/6G RAN}
\label{subsec:pipeline-arch}
\textit{SLED} connects field sensors to HPC over a private 5G/6G network, as shown in Fig.~\ref{fig:overview}. A field platform acts as the \textit{user equipment} (UE), carrying multiple sensor streams over the shared RAN. The gNB controls the UL and DL through its MAC scheduler, while the 5G/6G core manages sessions and forwards traffic through User Plane Functions (UPFs) toward the compute infrastructure. Following the O-RAN architecture, the gNB is split into a Distributed Unit (DU) and Centralized Unit (CU). The DU hosts the PHY, MAC, and RLC layers, including the MAC scheduler where \textit{SLED} performs its per-slot resource allocation. The CU hosts the control-plane and user-plane protocols and connects the RAN to the 5G core through the next generation(NG) interfaces.
\subsection{Sensor Slicing}
\label{subsec:SS}

With sensor slicing, we provide end-to-end isolation by virtualizing the UE, the RAN, and the core network. On the UE, each sensor is isolated with containers and Linux network namespaces, and its traffic is bound to a separate virtual network interface with a core-assigned IP address. To this end, the UE requests a separate Protocol Data Unit (PDU) session for each sensor, each of which is linked to a separate slice as identified by the Single Network Slice Selection Assistance Information (S-NSSAI). On the RAN, each PDU session is carried over its own data radio bearer (DRB), separating each sensor's traffic over the air and allowing per-sensor resource allocation within a shared UE. In the core network, each sensor slice is served by a separate UPF.

In the 5G RAN~\cite{3gpp_23501}, Layer 2 handles data buffering, segmentation, and scheduling. Within Layer 2, the Radio Link Control (RLC) layer buffers data for transmission, and the MAC layer schedules it. In sensor slicing, the Logical Channel ID (LCID) in the MAC helps map each slice instance to each data buffer. During the initial attach procedure, each sensor is given its own RLC buffer as well as its own LCID. During transmission, within a specific time slot, all of the PRBs are given to one sensor based on configured allocations. We modify the scheduling algorithm in the MAC layer to identify which sensor is allowed to transmit during this time slot. 
\begin{figure}[t!]
\centering
\includegraphics[
    width=0.95\columnwidth,
    clip
]{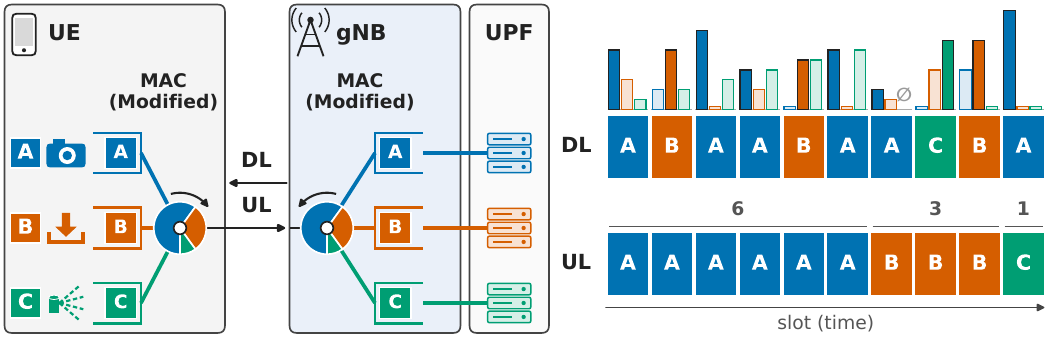}
\caption{Sensor Slicing algorithm with credit-based DL slot allocation (top) and deterministic UL slot ownership (bottom).}
\vspace{-0.25in}
\label{fig:algo}
\end{figure}
\subsubsection{Downlink}
In each DL slot, the PRBs are first allocated among the UEs by the gNB scheduler. Then, a credit-based weighted algorithm is used to determine which sensor within each UE is assigned the slot (Fig.~\ref{fig:algo}). Each sensor's slot assignments are spread evenly over time rather than grouped into consecutive blocks, thereby reducing scheduling delay while respecting the configured allocation. Initially, each sensor's credit is set to zero. Then, in each slot, the configured slice ratio of each sensor is added to its credit, but only for sensors with data to send. Sensors with no data are skipped, and their credits are not increased. The slot is assigned to the sensor with the highest credit for transmission. The sum of all sensor ratios is then subtracted from that sensor's credit. The updated credits are used as the starting values for the next slot.

\subsubsection{Uplink}
In each UL slot, the gNB scheduler first allocates PRBs among the UEs. A repeating cycle then determines which sensor within each UE owns the slot (Fig.~\ref{fig:algo}). Since the cycle is deterministic, the gNB and the UE can map slots to sensors independently, without per-slot signaling, which reduces control overhead. To this end, the gNB conveys the slice identifier, slice ratio, and LCID mapping of each sensor to the UE through a custom Slice Hint MAC Control Element. In the algorithm, the slice ratio of each sensor is first divided by the greatest common divisor (GCD) of all slice ratios, and each sensor is assigned the resulting number of consecutive UL slots within the cycle. For example, a 60/30/10 ratio for sensors A/B/C reduces to 6/3/1, so a 10-slot cycle allocates 6 slots to A, 3 to B, and 1 to C.

\subsection{Sensor-to-HPC Tunneling}
\label{subsec:tunneling}
\textit{SLED} extends the radio guarantee beyond the base station. It carries the slice identity through the core to the compute tier. Each slice uses a dedicated UPF as shown in Fig.~\ref{fig:overview} that forwards its traffic to the assigned inference tier through an authenticated SSH tunnel over TCP. The tier may be an edge GPU, a university HPC cluster, or Leadership Compute Facilities, rendering compute placement a per-slice configuration rather than a radio or application change. This makes compute placement transparent to the RAN, i.e., radio behavior remains unchanged across tiers, while application performance depends on the tier and path latency. We integrate DAQIRI~\cite{nvidia_daqiri} between the 5G core and the assigned GPU, following each slice's UPF. For each sensor stream, receiving many datagrams leads to multiple system calls on the host CPU. DAQIRI reassembles these datagrams into frames directly in pinned host CPU memory, transfers completed frames to the GPU, reducing CPU overhead associated with per-datagram system call processing.

\section{Implementation}
\label{sec:implementation}
In this section, we describe the implementation of the end-to-end sliced RAN pipeline. We first present the agentic workflow used to build and validate the slice-aware MAC scheduler. Then, we describe the sensor slicing algorithm used to enforce per-sensor resource allocations. 

\subsection{Agentic Build and Orchestration}
\begin{figure}[t!]
\centering
\includegraphics[
    width=0.90\columnwidth,
    trim={0 16.5cm 2cm 2cm},
    clip
]{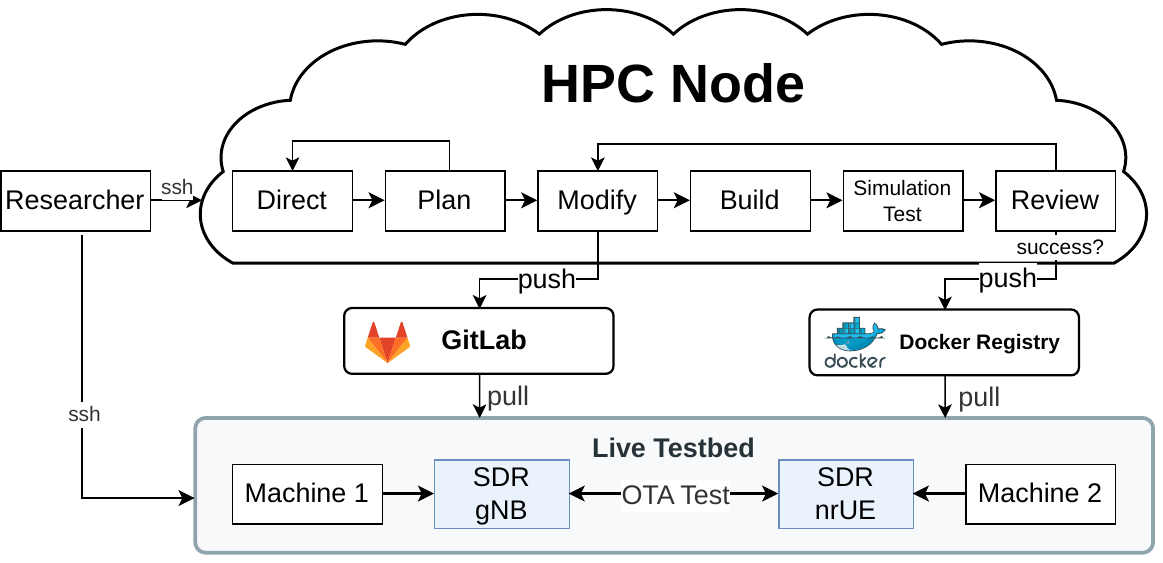}
\caption{Agentic RAN development harness with the automated modify-build-simulate loop on HPC and human-orchestrated OTA validation on a live 5G/6G RAN testbed.}
\vspace{-0.2in}
\label{fig:agentic_development}
\end{figure}

We build the sensor slicing feature into the 6G RAN by modifying the MAC layer of two O-RAN-compliant packages, OAI~\cite{oai} and OCUDU~\cite{ocudu}. These packages are compatible with the open-source, software-defined 5G NR UE (nrUE), which we modify for UL sensor scheduling. For the feature development (Fig.~\ref{fig:agentic_development}), we utilize an off-the-shelf agentic coding workflow with Claude Opus (versions 4.5--5) and Fable~5.1 models, adopting updates as they become available. We first implement the MAC modifications in the OAI gNB and nrUE, and then implement them in OCUDU.

\textbf{OAI}: First, we modify the OAI gNB and nrUE. For this, we run an agent inside a persistent tmux session on our institution's HPC machines, so we can assign it a task, disconnect, and return hours later to review the work. We require the agent to keep versioned git commits (v0.1, v0.2, \ldots, v8.5) and dated markdown logs of progress, successes, and failures. The progress logs give the agent context for its next run. The failure records keep the agent from repeating failed approaches. The success records let us later identify the prompts and decisions that steered the agent correctly.

\begin{table}[t!]
\caption{Agent token consumption by development activity.}
\label{table:token-activity}
\renewcommand{\arraystretch}{1.1}
\setlength{\tabcolsep}{3pt}
\setlength{\aboverulesep}{0pt}
\setlength{\belowrulesep}{0pt}
\resizebox{\columnwidth}{!}{%
\scriptsize
\begin{tabular}{@{}c|l r r r r r@{}}
\toprule
 & \textbf{Activity} & \textbf{Prompts} & \textbf{Input (M)$^{*}$} & \textbf{Output (M)} & \textbf{Total (M)} & \textbf{Share (\%)} \\
\midrule
\multirow{8}{*}{\rotatebox[origin=c]{90}{\textbf{OAI}}}
 & Debugging      & 386 &   997.0 & 3.79 & 1,000.7 & 70.1 \\
 & Implementation &  39 &   137.2 & 0.27 &   137.4 &  9.6 \\
 & Testing        &  77 &   106.7 & 0.17 &   106.9 &  7.5 \\
 & Building       &  62 &    78.5 & 0.20 &    78.7 &  5.5 \\
 & Infrastructure &  74 &    52.1 & 0.18 &    52.3 &  3.7 \\
 & Learning       & 199 &    46.9 & 0.34 &    47.2 &  3.3 \\
 & Meta           &  76 &     5.3 & 0.03 &     5.3 &  0.4 \\
\cmidrule{2-7}
 & \textbf{Total} & \textbf{913} & \textbf{1,423.6} & \textbf{4.97} & \textbf{1,428.5} & \textbf{100} \\
\midrule
\multirow{8}{*}{\rotatebox[origin=c]{90}{\textbf{OCUDU}}}
 & Debugging      &   5 &  49.4 & 0.08 &  49.5 &  7.7 \\
 & Implementation &   7 & 156.7 & 0.16 & 156.8 & 24.4 \\
 & Testing        &  15 & 228.9 & 0.19 & 229.1 & 35.6 \\
 & Building       &  15 & 174.3 & 0.20 & 174.5 & 27.1 \\
 & Infrastructure &   9 &  20.3 & 0.03 &  20.4 &  3.2 \\
 & Learning       &   0 &   0.0 & 0.00 &   0.0 &  0.0 \\
 & Meta           &   2 &  12.6 & 0.01 &  12.6 &  2.0 \\
\cmidrule{2-7}
 & \textbf{Total} & \textbf{53} & \textbf{642.3} & \textbf{0.67} & \textbf{643.0} & \textbf{100} \\
\bottomrule
\end{tabular}}
\vspace{-0.15in}
\end{table}
We begin with an unmodified clone of OAI's develop branch~\cite{oai}. We provide the agent with reference materials on existing open-source slicing implementations, including srsRAN and ORANSlice, to study before developing the modifications in our OAI clone. Next, we provide it with high-level instructions on the scheduler modifications and define the success criteria. We instruct it to implement the changes, rebuild the code in a containerized environment, and test the changes in simulation. Initially, we provide detailed testing instructions, e.g., ping for connectivity and iperf for per-tunnel throughput. Once these pass, we incorporate these instructions into reusable skills and hooks to help automate subsequent tests. We also utilize subagents to perform independent tasks in parallel, shortening development time.

Versions (v0.1–v0.4) focused on DL sensor slicing enforcement, where we modified the gNB to map each sensor’s logical channel to its assigned slice, track buffered channels, and enforce resource allocations per the slice policy file. The first DL slice enforcement was verified about 5.5 h after development began. Afterward, we extended the implementation to UL sensor slicing, which required changes to both the gNB and the UE. In the final releases (v8.3–v8.5), we consistently measured each sensor’s throughput share close to its configured slice allocation in both UL and DL. We then containerized the final release for human-driven OTA deployment and testing. \textbf{Over 6.5 weeks}, the agent produced 122 commits across 29 OAI files and consumed approximately 1.43B tokens.

\textbf{OCUDU:} Next, we implement sensor slicing on a second O-RAN compliant package, OCUDU~\cite{ocudu}. To speed up the development process, we use the validated OAI implementation of sensor slicing as the reference, with the modified OAI source code serving as the code reference and the OAI development logs identifying the approaches that had failed. During development, we utilize a coordinator agent that distributes the work among parallel subagents. These agents are each assigned a separate task: one agent studies our SL implementation in the OAI codebase and implements the feature in OCUDU, other agents build the 5G core and the original OAI stack for comparison, and another agent integrates the OCUDU gNB with the OAI nrUE in the RF simulator. A separate reviewer agent checks the resulting implementations and builds. The first OCUDU DL sensor slicing enforcement passed its unit tests 1.2\,h after development began, and the end-to-end SL implementation, including UL, was completed and tested in simulation \textbf{within 2\,h\,40\,min}.

Token consumption for the development activities is listed in Table~\ref{table:token-activity}. In implementation, we modify the RAN source code to implement the sensor slicing feature; in debugging, we review and correct the agent's decisions; in building, we compile and containerize the gNB and UE; in testing, we validate per-slice throughput in the RF simulator; in infrastructure, we manage the HPC environment, git history, and images; in learning, we study the NR specifications and the RAN code base; and in meta, we maintain commits and progress logs.

The impact of a validated reference implementation on agentic development is substantial, as shown in Table~\ref{table:token-activity}. Compared with OAI, OCUDU development required 400$\times$ less time (2\,h\,40\,min vs.\ 6.5~weeks), 228$\times$ fewer human prompts (4 vs.\ 913), and less than half the tokens (0.64B vs.\ 1.43B). Debugging accounted for only 7.7\% of OCUDU tokens, compared with 70.1\% for OAI. Next, we describe the sensor slicing modifications in the gNB, UE, and core network.

\subsection{Sensor Slicing Implementation}
\label{sec:impl-ss}

\begin{figure}[t!]
\centering
\includegraphics[
    width=0.85\columnwidth,
    clip
]{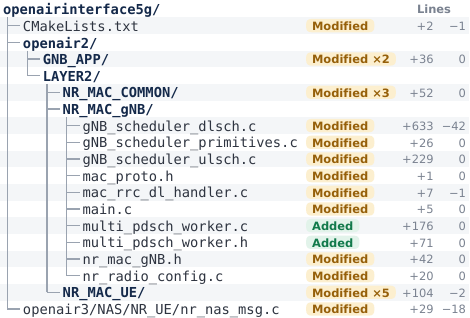}
\caption{OAI file modifications}
\vspace{-0.2in}
\label{fig:oai_file_modifications}
\end{figure}

For sensor slicing, we modify the gNB and UE source files and update the configuration file in the core network. \textbf{gNB:} On the gNB, we modify the downlink scheduler at the MAC layer to decide which sensor gets each slot, and the uplink scheduler to give each slot to the sensor that owns it. \textbf{UE:} We modify the UE MAC layer, so that the UE sends each sensor's data in the same uplink slots that the gNB assigns to that sensor. Outside of the MAC layer, we make one fix in the UE's NAS layer. Each sensor requires its own Protocol Data Unit (PDU) session, which is a data path to the core with its own slice identifier. To set up a PDU session, the core sends the UE a control message, called the PDU Session Establishment Accept. The UE expects these control messages to arrive in order. When a UE requests multiple PDU sessions at the same time, one per sensor, these control messages can arrive at the UE out of order, and the legacy OAI rejects the late messages. We update the NAS layer such that the UE accepts these control messages in any order, so that each sensor gets its own PDU session. \textbf{Core:} On the core network (Open5GS), we update the configuration file such that each sensor's slice has its own UPF.

\section{Evaluations}
\label{sec:evaluation}
\begin{figure}[t!]
\centering
\includegraphics[width=0.8\columnwidth]{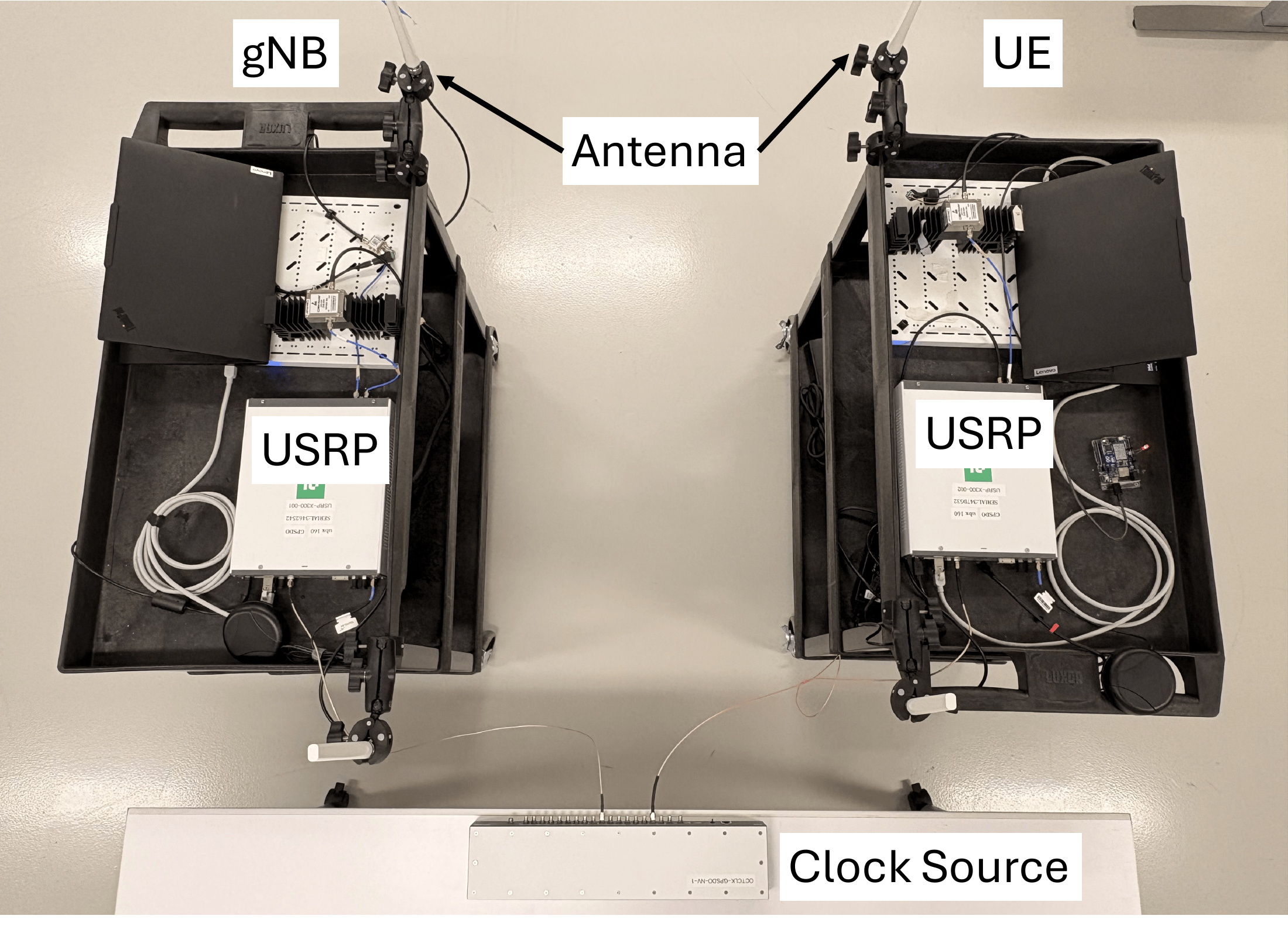}
\caption{Indoor OTA Testbed (top view)} 
\label{fig:setup_indoor}
\vspace{-0.2in} 
\end{figure}

In this section, we evaluate \textit{SLED} in two complementary settings: an indoor OTA testbed(Fig.~\ref{fig:setup_indoor}) and a lossless radio frequency (RF) simulator. The OTA experiments validate the full sensor-to-HPC pipeline and per-slice isolation capabilities of sensor slicing under realistic radio conditions, while the RF-simulator study allows the scheduling behavior to be scaled and isolated by removing channel loss. The experiment setup is described in Sec.~\ref{subsec:eval-setup}. In Sec.~\ref{subsec:eval-internal}, \textit{SLED} is evaluated across different compute tiers, radio resource allocation policies, and numbers of sensor slices (SLs) per UE. In Sec.~\ref{subsec:eval-external}, \textit{SLED} is compared with ORANSlice and standard 5G network slicing. Finally, in Sec.~\ref{subsec:eval-hpc}, we show \textit{SLED} supporting HPC-grade AI workloads from sensors to compute.

\subsection{Experiment Setup}
\label{subsec:eval-setup}
The OTA testbed (Fig.~\ref{fig:setup_indoor}) consists of a gNB and a UE, each equipped with a USRP X300 radio and positioned $2$\,m apart. SLED is built on modified 5G OAI~\cite{oai} RAN, UE, and Open5GS~\cite{open5gs} core packages. The radios use omnidirectional antennas operating at band n48 ($3.6$\,GHz), with a bandwidth of $40$\,MHz, subcarrier spacing of $30$\,kHz, and $106$ resource blocks. A TDD pattern of \texttt{DDDDDSUUUU} ($5$ DL, $4$ UL) is used. The gNB and UE are hosted on Intel Xeon Core Ultra 7 processors. We also use OAI RF simulator~\cite{oai} to evaluate scalability under lossless radio channel conditions. All evaluations are performed using OAI gNB and OAI nrUE. We also evaluate the scalability of the packages using both OAI gNB vs.~OCUDU gNB with OAI nrUE. The testbed provides three compute tiers for sensor traffic: (1) Edge device located at the base station (Edge) with an NVIDIA A2000 GPU, (2) Campus HPC cluster (Campus) with an NVIDIA V100 GPU, and (3) a DOE leadership computing facility (LCF) -- Perlmutter from NERSC -- with NVIDIA A100 GPUs. Each experiment is repeated over 50 trials. 

\begin{table}[t!]
\caption{Slice workload characteristics.}
\label{table:slices}
\renewcommand{\arraystretch}{1.1}
\setlength{\tabcolsep}{3pt}
\resizebox{\columnwidth}{!}{%
\scriptsize
\begin{tabular}{@{}l l l r r r@{}}
\toprule
\textbf{Slice} & \textbf{Application} & \textbf{Transport} & \textbf{Packet} & \textbf{Data Rate} & \textbf{5QI~\cite{3gpp_23501}} \\
 & & \textbf{Protocol} & \textbf{Size} & & \\
\midrule
eMBB-VA  & Video Analytics (JPEG)   & UDP & 1,400\,B & 15 Mbps         & 2  \\
eMBB-BT  & Bulk Transfer            & TCP &  1,400\,B & 50 Mbps            & 9  \\
eMBB-CL  & Continual Learning       & UDP & 1,400\,B & 18 Mbps           & 4  \\
eMBB-IQ  & Raw Radio (IQ) Streaming & TCP & 360\,KB  & 25 Mbps           & 9  \\
eMBB-LD  & LiDAR Transfer           & UDP & 1,400\,B & 8 Mbps      & 3  \\
URLLC-RT & Real-Time Control        & UDP & 975\,B   & 0.4 Mbps & 84 \\
mMTC-MSG & Message Log              & UDP & 1,400\,B & 5 Mbps       & 8  \\
\bottomrule
\end{tabular}}
\vspace{-0.15in}
\end{table}

\subsection{Sensor Workloads}
Throughout the evaluations, we employ seven workload configurations (Table~\ref{table:slices}) designed to represent realistic IoT and AI-IoT applications: \textbf{(i) Enhanced mobile broadband visual analytics (eMBB-VA):} JPEG-encoded $640\times480$ frame streaming over two parallel UDP flows at 52 frames/s each to a server running YOLOv11 object detection~\cite{ultralytics_yolo}, which returns detections. \textbf{(ii) Ultra-reliable low-latency robot telemetry (URLLC-RT):} 975-B payloads sent at 50 Hz, emulating robot telemetry with a 30-ms latency requirement corresponding to 3GPP 5QI~\cite{3gpp_23501} 84. \textbf{(iii) eMBB bulk transfer (eMBB-BT):} Long-lived, backlogged TCP traffic using iperf3. \textbf{(iv) eMBB LiDAR (eMBB-LD):} Continuous, multi-megabit stream from a single sensor, where sustaining a high data rate is the main requirement~\cite{itu_m2083}. \textbf{(v) Massive machine-type communication (mMTC-MSG):} Small, delay-tolerant messages from many low-power sensors, intended for a robot to aggregate and relay to the network~\cite{itu_m2083}. \textbf{(vi) eMBB continual learning (eMBB-CL):} eMBB-VA frame streaming to the Campus HPC compute tier, where YOLOv12x~\cite{tian2026yolov12} annotates each frame and the annotations train a smaller YOLOv12n model online, keeping the model robust to data drift as field conditions change. Annotated frames and training performance are returned over the DL. \textbf{(vii) eMBB IQ sensing (eMBB-IQ):} Raw IQ data streaming in SIGMF format~\cite{hilburn2018sigmf} via UE’s UL slots. IQ samples are recorded in a queue, and transmitted during consecutive ULs to the LCF compute tier. The DL slots include signal quality reports with frequency offset and spectrum occupancy information.

All workloads except eMBB-BT and eMBB-IQ use raw UDP without retransmission, exposing contention through packet or frame loss rather than transport recovery. Frames and messages are fragmented into MTU-sized datagrams, and the server tracks whether each is fully reconstructed, while telemetry uses sequence numbers and timestamps to measure loss and round-trip latency. Each server runs as a microservice in the core network namespace, and each sender is bound to its slice tunnel to ensure traffic traverses the intended radio slice. We record per-slice delivered UL and DL throughput, packet delivery rate, and application round-trip time.
\subsection{Sensor Slicing Evaluation and Scalability}
\label{subsec:eval-internal}
We first evaluate sensor slicing in \textit{SLED} in two stages: Radio resource allocation across sensor streams and scaling the number of SLs, using the RF simulator.
\begin{figure}[t!]
\centering
\includegraphics[width=\columnwidth]{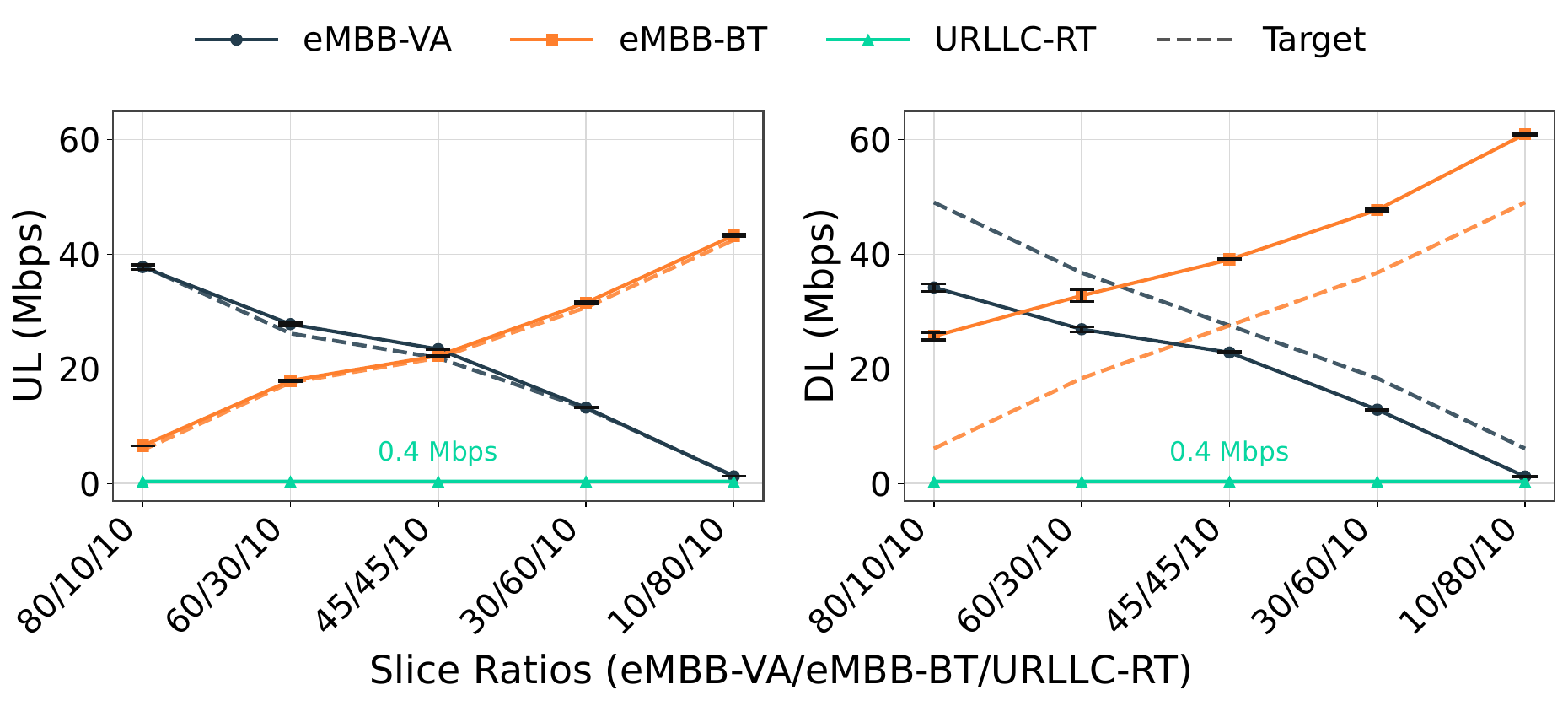}
\caption{Per-slice OTA uplink (left) and downlink (right) throughput with $95\%$ confidence intervals.}
\vspace{-0.2in}
\label{fig:eval-minratio}
\end{figure}
\subsubsection{Resource Allocation Policy}
\label{subsubsec:eval-minratio}
To quantify how the allocated resource for a SL affects its delivered throughput, we vary the minimum PRB ratio between the eMBB-VA and eMBB-BT slices across five policies (from 80/10/10 to 10/80/10) on the OTA testbed. URLLC-RT is held at a fixed 10\% ratio in all configurations to meet the 30 ms latency requirement of 3GPP 5QI 84. With 0.5 ms slots and a 5DL/4UL TDD pattern, this allocation provides transmission opportunities every 10 ms. Combined with approximately 15 ms of RAN latency and server-side processing, the resulting end-to-end latency is 28--29 ms, making 10\% the minimum allocation under our scheduling policy.

The results, as shown in Fig. \ref{fig:eval-minratio}, show a clear correlation between PRB allocation and throughput in the UL. As each eMBB slice is assigned a larger share, its throughput increases consistently. eMBB-VA decreases by 36.49 Mbps from 80\% to 10\%, while eMBB-BT rises by 36.67 Mbps over the complementary shares. URLLC-RT holds a constant rate throughout. The DL preserves this ordering at lower throughput, with eMBB-VA falling by 32.97 Mbps while eMBB-BT remains near 41 Mbps, as the minimum PRB ratio is a guarantee, not a cap. eMBB-VA carries inference replies and does not offer enough DL traffic to consume its share, so the scheduler redistributes the unused PRBs to the eMBB-BT slice. UL throughput scales with assigned PRBs, while DL throughput follows offered load without violating guarantees. Overall, sensor slicing provides controlled and predictable resource allocation within a single UE.

\begin{figure}[t!]
\centering
  \includegraphics[width=0.90\columnwidth]{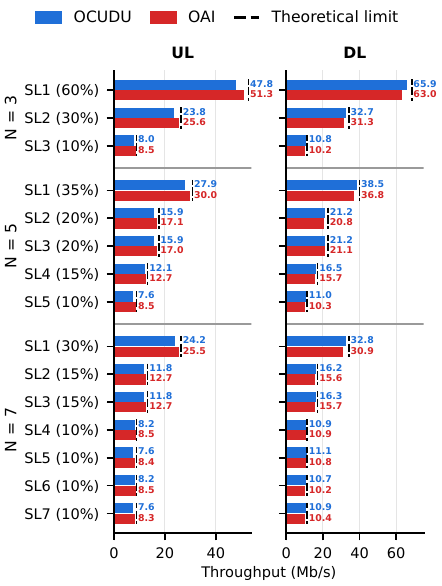}
\caption{Per-slice allocation (Alc, \%) and achieved UL/DL throughput (Mbps) for a UE with $N=\{3,5,7\}$ SLs.}
\vspace{-0.23in}
\label{fig:scal}
\end{figure}
\subsubsection{Scalability Across Sensor Slices}
\label{subsubsec:eval-scaling}
To evaluate the scalability performance, we configure a single UE with three, five, and seven concurrent slices in the RF simulator, with both the OAI gNB and the OCUDU gNB, and OAI nrUE. A saturating UDP iperf3 stream is transmitted simultaneously on each slice for 120\,s in each direction, repeated 50~times per configuration. The theoretical throughput of each slice is its allocation multiplied by the theoretical cell capacity (88.0~Mb/s UL and 114.2~Mb/s DL, from the 3GPP transport block size). As shown in Fig.~\ref{fig:scal}, per-sensor isolation is preserved for both gNBs as the number of configured slices grows. In all cases, the measured throughput of each slice is on average within 0.9\% (OAI) and 1.4\% (OCUDU) of its expected throughput according to the slice ratio.

\subsection{Baseline Comparisons against State-of-the-Art}
\label{subsec:eval-external}
In this section, we compare \textit{SLED} against OAI~\cite{oai} network slicing and ORANSlice~\cite{cheng2024oranslice} using the RF simulator. 
\subsubsection{Two-Slice~Downlink~Isolation} 
To account for ORANSlice's two-slice DL limitation, we compare network slicing, ORANSlice, and sensor slicing using eMBB-VA and eMBB-BT workloads. For network slicing, we utilize a single network slice for a single UE with 100\% radio resource allocation for both workloads. For ORANSlice and sensor slicing, we configure a single UE with two SLs (one per workload) with a 70/30 resource allocation. Both workloads are transmitted via UDP DL. eMBB-VA is kept at a constant rate of 52 fps while eMBB-BT data rate is increased from 0 to 60 Mbps in 5 Mbps increments every 8\,s.
\begin{figure}[t!]
\centering
\includegraphics[width=0.8\columnwidth]{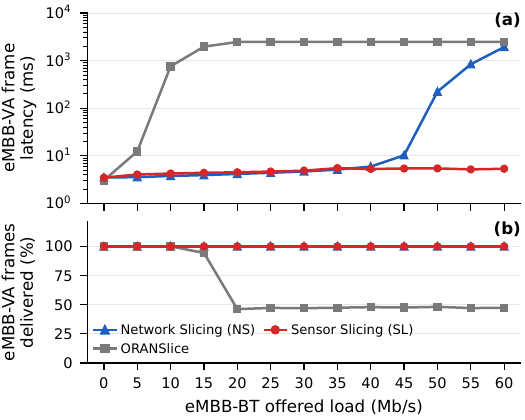}
\caption{Two-slice downlink latency and delivery ratio for Sensor Slicing vs.~State-of-the-Art (DL, 70/30, 52~fps).}
\label{fig:sl_s_os_v_ns}
\end{figure}
The results, as shown in Fig. ~\ref{fig:sl_s_os_v_ns}, show a clear difference in how each solution isolates eMBB-VA as eMBB-BT load increases. With sensor slicing, the eMBB-VA latency remains between 3.5 and 5.5~ms and 100\% of frames are delivered.
In network slicing, the eMBB-VA latency increases from 3.5 ms--4.7 ms to 223 ms at 50 Mbps and 1.93 s at 60 Mbps because both streams share a single slice, and the MAC scheduler lacks the granularity to prioritize eMBB-VA over eMBB-BT. Moreover, in ORANSlice, eMBB-VA latency increases from 3 ms at 0 Mbps of eMBB-BT load to 733 ms at 10 Mbps and to 2.46 s after 20 Mbps load. At the same time, only 46\%--48\% of frames are delivered. This is because ORANSlice alternates the slots between the two slices, and limits the PRBs of each slice within its slot. Thus, eMBB-VA receives only about half of its 70\% allocated radio resource.

\subsubsection{Multi-Sensor Uplink/Downlink Isolation}
\begin{figure}[t!]
\centering
\includegraphics[width=0.90 \columnwidth]{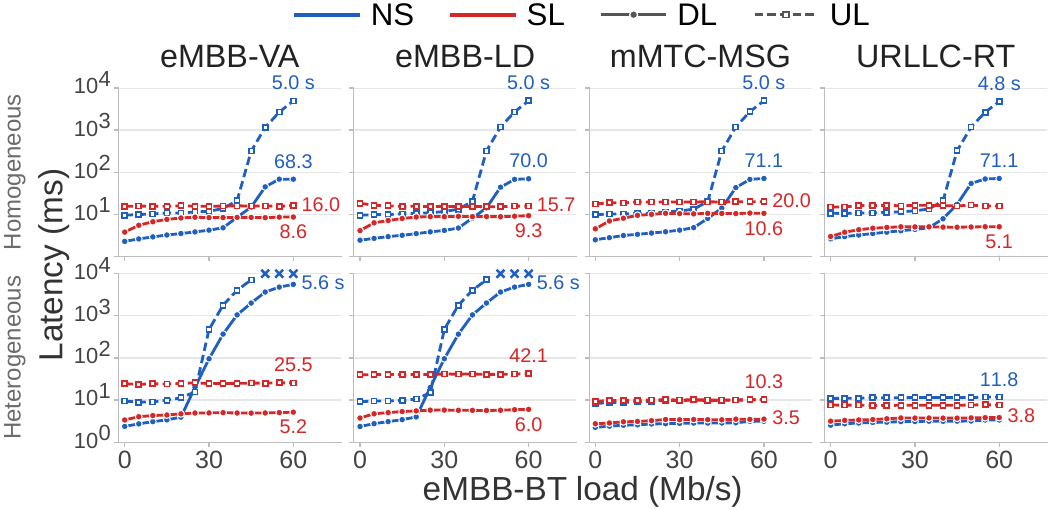}
\caption{Latency of each application under network slicing (NS) and sensor slicing (SL) as the eMBB-BT load increases.}
\vspace{-0.25in}
\label{fig:ss_vs_ns}
\end{figure}

Next, we scale SLs for workloads involving both UL and DL transmission and compare sensor slicing with network slicing in a five-sensor scenario with two UEs (Fig.~\ref{fig:ss_vs_ns}) in the RF simulator. ORANSlice is excluded, as it supports at most two slices per UE and no UL isolation. The following five sensor streams are used: eMBB-VA, eMBB-BT, eMBB-LD, URLLC-RT, and mMTC-MSG). The eMBB-VA, eMBB-LD, mMTC-MSG, and URLLC-RT streams are transmitted at a fixed rates of 12.3, 7.0, 5.3, and 0.39 Mbps in each direction. These rates are set proportional to the sensor slice ratios, such that each stream remains within its allocated share and any degradation is caused only by the eMBB-BT load. We vary the eMBB-BT offered load from 0 to 60 Mbps and measure the throughput and latency of the remaining workloads. We conduct the experiments in two ways, depending on how the two traffic classes are distributed among the UEs. 
\textit{Homogeneous UEs:} In this setup, both UEs carry the same set of sensors. In network slicing, we create a single network slice instance shared by both UEs that carries all sensor workloads from both UEs. In sensor slicing, we create separate SL instances for each sensor type, shared by traffic of the same sensor type across both UEs. In network slicing, as the eMBB-BT load increases from 0 to 60 Mbps, the latency of the other workloads increases from $\sim$2.3 ms to $\sim$71 ms. This occurs because all workloads need to share the same network slice, and as the eMBB-BT throughput rises, its packets fill the buffer. In contrast, sensor slicing dedicates a separate SL to each sensor type. As a result, each sensor traffic stream is isolated from increasing traffic from another sensor workload.

\textit{Heterogeneous UEs:} In this setup, each UE carries sensors from a single class: UE1 carries the eMBB-class sensors, and UE2 carries URLLC-RT and mMTC-MSG. In network slicing, we assign one network slice per UE based on its traffic class. In sensor slicing, each sensor within a UE is assigned its own sensor slice. For network slicing, URLLC-RT and mMTC-MSG are unaffected as the eMBB-BT load increases, since they are on a different UE and slice instance. However, eMBB-VA and eMBB-LD share the same network slice as eMBB-BT. As the eMBB-BT load increases from 0 to 60 Mbps, its latency increases from 2.4 ms to 5.6 s, with an increase of more than 2,300-fold relative to baseline. With SL, each sensor has its own separate sensor slice within the UE. As a result, eMBB-VA and eMBB-LD maintain stable latency and throughput.


\subsection{End-to-end OTA Sensor-to-HPC Evaluations}
\label{subsec:eval-hpc}
Next, we evaluate SLED over an end-to-end multi-tier system to quantify the impacts of compute placement and AI/SI workloads in the OTA testbed. 
\subsubsection{Compute Placement and End-to-end Isolation}
\label{subsubsec:eval-hero}
To observe the impact of compute placement, we couple the eMBB-VA stream with each compute tier (Edge, Campus, and LCF) in three separate experiments. The remaining eMBB-BT and URLLC-RT slices terminate at the Edge in every configuration. To test the effect of compute placement, the RAN is configured to allocate the radio resources within a $60/30/10$ policy, where eMBB-VA, eMBB-BT, and URLLC-RT sensor streams are guaranteed $60\%$, $30\%$, and $10\%$ of the resources, respectively. 
This emulates a scenario where sensor streams from the same source must flow to various compute nodes for isolated service requirements. Each experiment is conducted for $120$\,s.
\begin{figure}[t!]
\centering
\includegraphics[width=0.90\columnwidth]{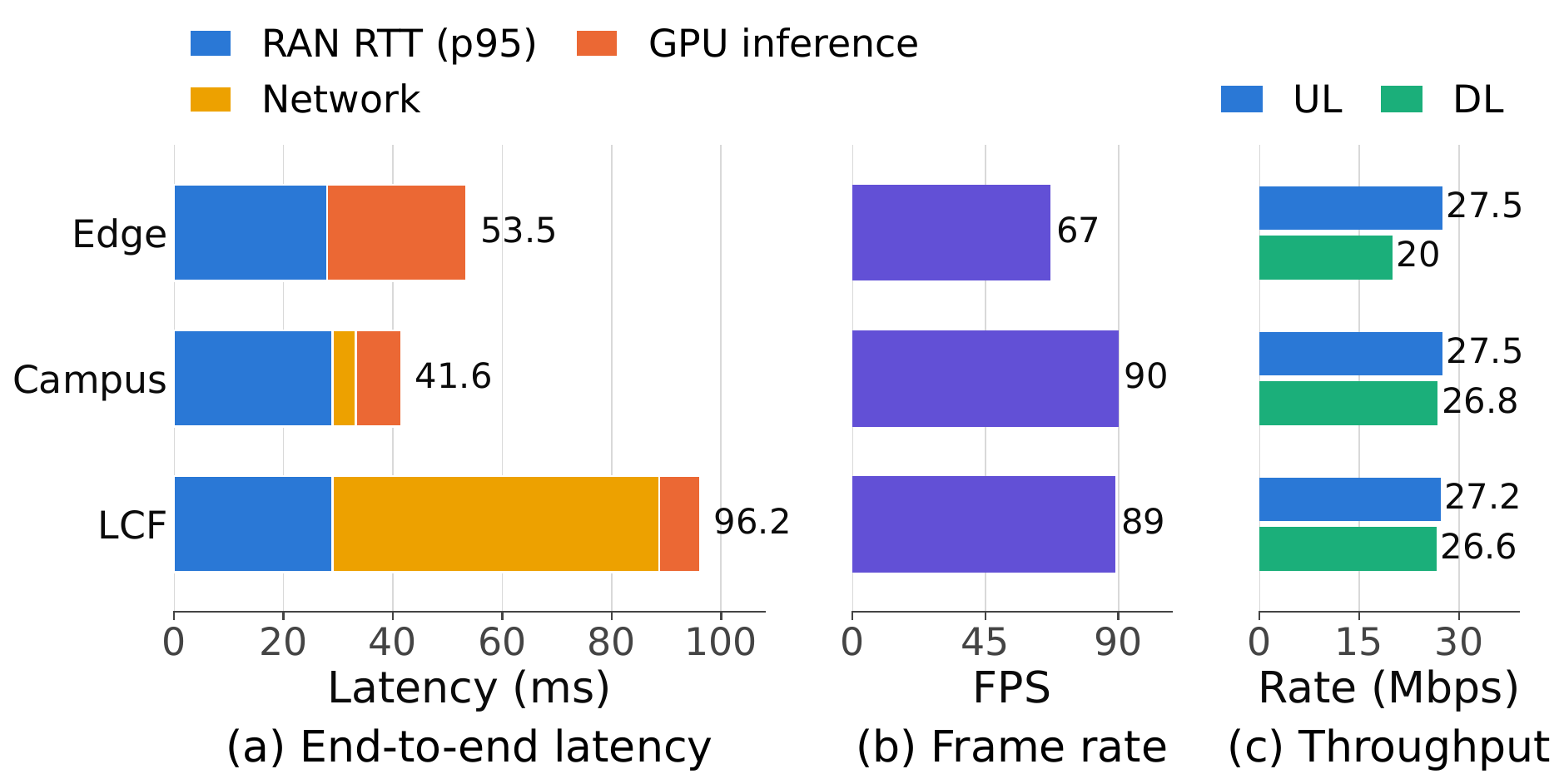}
\caption{Compute placement impact on eMBB-VA latency, frame rate, and throughput.}
\label{fig:va-compute}
\vspace{-0.20in}
\end{figure}
\textbf{Compute Placement}: The delivered eMBB-VA latency for each compute, processed frame rate, and throughput, and the resulting processed frame rates are shown in Fig~\ref{fig:va-compute}. The results reveal interesting trade-offs in the sensor-to-HPC path. RTT is computed as a sum of the RAN user-plane latency, network latency, and GPU inference time. RAN user-plane RTT remains stable (between $28$--$29$ ms for all cases), while the application-level RTT changes because of the tradeoff between GPU inference latency and the additional network path to the remote compute tier. Edge has the lowest network latency but the highest inference latency ($25.5$ ms per frame), while Campus and LCF provide faster inference ($8.4$ ms and $7.5$ ms) at the cost of additional network latency ($4.2$ ms and $59.7$ ms). As a result, Edge, Campus, and LCF achieve $67$, $90$, and $89$ fps, respectively. In this scenario, Campus strikes a balance between inference time and network delay, resulting in the highest frame rate. Interestingly, LCF significantly outperforms Edge despite the longer network latency because of its $71\%$ faster compute. Notably, the tradeoff between network latency and inference cost may change depending on topology and workload requirements. \textit{SLED} can dynamically support this tradeoff by guaranteeing end-to-end isolated data tunnels that can be flexibly coupled with different compute tiers based on sensor service requirements.

\textbf{Per-sensor Isolation:} The impacts of eMBB-VA compute placement on eMBB-BT and URLLC-RT reveal important cross-traffic tradeoffs. First, \textit{SLED} successfully isolates the vulnerable URLLC-RT slice so that the full $400$\,kbps allocation is achieved at constant $28$--$29$\,ms p95 RTT in each case. This is possible with sensor slicing, as each sensor stream has its own queue and every slot is owned by one single slice at a time. Similar isolation is observed for eMBB-BT when it does not share compute placement with eMBB-VA. 

When eMBB-BT and eMBB-VA share the same Edge GPU, we observe fluctuations in DL throughput of both flows. As shown in Fig~\ref{fig:va-compute}, eMBB-VA DL throughput at Edge is $25\%$ lower than at Campus. Higher inference time at Edge results in only $72\%$ of frames being processed. Interestingly, eMBB-BT throughput increases by $13\%$ with Edge compared to Campus due to a cross-layer interaction: When eMBB-VA underutilizes its guaranteed radio slots, sensor slicing reallocates them to the elastic eMBB-BT traffic, increasing throughput. This highlights a cross-layer design opportunity for \textit{SLED}.

\subsubsection{Real-world HPC-Grade Workloads}
To show that \textit{SLED} supports real-world sensor-to-HPC AI/SI workloads, we use eMBB-CL, eMBB-IQ, and URLLC-RT workloads. The allocation policy is defined to be $40\%$, $50\%$, and $10\%$ of the radio resources, respectively.
\begin{table}[t!]
\centering
\footnotesize
\captionsetup{font=footnotesize, textfont=it, justification=raggedright, singlelinecheck=false, skip=3pt}
\caption{HPC-grade workloads at a 40/50/10 allocation.}
\label{tab:Realworldworkflows}
\renewcommand{\arraystretch}{1.0}
\setlength{\tabcolsep}{3pt}
\begin{tabular*}{\columnwidth}{@{\extracolsep{\fill}}llrrrr@{}}
\toprule
\textbf{Slice} & \textbf{Compute} & \textbf{Required} & \textbf{Measured} & \textbf{Delivered} & \textbf{UL} \\
 &\textbf{Location} & & & \textbf{(\%)} & \textbf{(Mbps)} \\
\midrule
eMBB-CL  & Campus HPC & 60\,fps     & 60.3\,fps & 99.9 & 18.1 \\
eMBB-IQ  & LCF        & 0\% loss    & 0\% loss  & 100  & 25.6 \\
URLLC-RT & Edge       & $\le$30\,ms & 29\,ms    & 100  & 0.4 \\
\bottomrule
\end{tabular*}\vspace{-0.22in}
\end{table}

As shown in Table~\ref{tab:Realworldworkflows}, \textit{SLED} supports two different real-world workloads to different endpoints while maintaining per-slice isolation. The DOE LCF processes the eMBB-IQ data, while the Campus HPC cluster runs the eMBB-CL workflow. eMBB-CL streams $60$ fps at $18.1$\,Mbps, using approximately $92\%$ of its ideal UL throughput. It delivers $99\%$ of frames, with a latency of $81$ ms, while the YOLOv12x model learns from incoming data. Similarly, eMBB-IQ uses its full UL allocation, reaching $25.6$ Mbps, and delivers all recordings with a latency of $285$ ms, of which $165$ ms is spent transferring data from the RAN gNB to the DOE LCF. Meanwhile, URLLC-RT maintains $0.4$ Mbps with $100\%$ delivery and a $29$ ms latency despite the competing workloads. This shows \textit{SLED} facilitating two heavy real-world eMBB workloads with different endpoints while providing persistent isolation and guarantees to the URLLC Slice.
\section{Conclusion}
\label{sec:conclusion}
We present \textit{SLED}, an end-to-end framework that transitions O-RAN resource management from device-centric allocation to fine-grained, sensor-specific slicing. By virtually coupling the sensor, RAN, Core, and compute tiers, SLED enables deterministic isolation of heterogeneous streams. At its core, sensor slicing maps each sensor stream to its own slice in both uplink and downlink and enforces its configured resource allocation, even when streams share a UE. Using agentic coding techniques, we implement sensor slicing in two O-RAN packages, OAI and OCUDU, where the validated OAI implementation reduces the OCUDU development from weeks to hours. Experimental results demonstrate that slice isolation holds as the number of streams grows and as computation moves across edge, campus, and remote HPC resources. Furthermore, we demonstrate that SLED allows for flexible coupling of sensor streams to diverse compute endpoints without requiring RAN reconfiguration while maintaining predictable end-to-end performance. \textit{SLED} provides a scalable foundation for next-gen AI/SI-native, sensor-to-HPC pipelines in 6G networks.

\section{Acknowledgments}
This work was supported in part by the US Department of Energy grant DE-SC0025541 and GENESIS MISSION grant DE-SC0026804. The research used resources of the National Energy Research Scientific Computing Center (NERSC), a Department of Energy User Facility, using NERSC award ERCAP0039181; and resources of the Holland Computing Center of the University of Nebraska, which receives support from the UNL Office of Research and Innovation, and the Nebraska Research Initiative.

AI models were used to develop the sensor slicing implementations through the agentic workflows described in Sec.~\ref{sec:implementation} and to assist with experiment scripts and plotting, and all content was reviewed and verified by the authors.
\begingroup
\bibliographystyle{IEEEtran}
\bibliography{references_clean}

@article{mach2017mobile,
  title={Mobile edge computing: A survey on architecture and computation offloading},
  author={Mach, Pavel and Becvar, Zdenek},
  journal={IEEE Communications Surveys \& Tutorials},
  volume={19},
  number={3},
  pages={1628--1656},
  year={2017},
  publisher={IEEE}
}

@inproceedings{kurafeeva2025xgfabric,
  author    = {Kurafeeva, Liubov and Subedi, Alan and Hartung, Ryan and Fay, Michael and Biswas, Avhishek and Jha, Shantenu and others},
  title     = {{xGFabric}: Coupling Sensor Networks and {HPC} Facilities With Private {5G} Wireless Networks for Real-Time Digital Agriculture},
  booktitle = {Proceedings of the SC '25 Workshops},
  year      = {2025}
}

@article{albiero2022swarm,
  title={Swarm robots in mechanized agricultural operations: A review about challenges for research},
  author={Albiero, Daniel and Garcia, Angel Pontin and Umezu, Claudio Kiyoshi and de Paulo, Rodrigo Leme},
  journal={Computers and Electronics in Agriculture},
  volume={193},
  pages={106608},
  year={2022},
  publisher={Elsevier}
}

@article{popovski20185g,
  title={{5G} Wireless Network Slicing for {eMBB}, {URLLC}, and {mMTC}: A Communication-Theoretic View},
  author={Popovski, Petar and Trillingsgaard, Kasper Fl{\o}e and Simeone, Osvaldo and Durisi, Giuseppe},
  journal={IEEE Access},
  volume={6},
  pages={55765--55779},
  year={2018},
  publisher={IEEE}
}

@techreport{3gpp2026,
  author      = {{ETSI}},
  title       = {{5G; NR; NR and NG-RAN Overall Description; Stage-2}},
  institution = {European Telecommunications Standards Institute},
  number      = {ETSI TS 138 300 V19.3.0 (Rel. 19)},
  year        = {2026},
  month       = aug,
}

@inproceedings{cheng2024oranslice,
  author    = {Cheng, Hai and D'Oro, Salvatore and Gangula, Rajeev and Velumani, Sakthivel and Villa, Davide and Bonati, Leonardo and Polese, Michele and Arrobo, Gabriel and Maciocco, Christian and Melodia, Tommaso},
  title     = {{ORANSlice}: An Open-Source {5G} Network Slicing Platform for {O-RAN}},
  booktitle = {Proc. ACM MobiCom},
  year      = {2024}
}

@misc{oai,
  author       = {{OpenAirInterface Software Alliance}},
  title        = {{OpenAirInterface}: Open-Source 5G {NR} Software},
  howpublished = {\url{https://gitlab.eurecom.fr/oai/openairinterface5g}},
  note         = {Develop branch, commit \texttt{40e61d59}, weekly tag
                  \texttt{2026.w07}},
  year         = {2026}
}

@misc{ocudu,
  author = {{OCUDU Ecosystem Foundation}}, title = {{OCUDU}: Open-Source {5G} {CU/DU} Project},
  howpublished = {\url{https://ocudu.org}}, year = {2026}}

@article{polese2023coloran,
  author  = {Polese, Michele and Bonati, Leonardo and D'Oro, Salvatore and Basagni, Stefano and Melodia, Tommaso},
  title   = {{ColO-RAN}: Developing Machine Learning-Based {xApps} for Open {RAN} Closed-Loop Control on Programmable Experimental Platforms},
  journal = {IEEE Transactions on Mobile Computing},
  volume  = {22},
  number  = {10},
  pages   = {5787--5800},
  year    = {2023}
}

@inproceedings{dorcheh2025dora,
  author  = {Ebrahimi Dorcheh, Alireza and Seyfi, Tolunay and Afghah, Fatemeh},
  title   = {{DORA}: Dynamic {O-RAN} Resource Allocation for Multi-Slice {5G} Networks},
  booktitle = {IEEE Middle East Conference on Communications and Networking},
  year    = {2025},
}

@inproceedings{barker2025real,
  author  = {Barker, Ryan and Ebrahimi Dorcheh, Alireza and Seyfi, Tolunay and Afghah, Fatemeh},
  title   = {{REAL}: Reinforcement Learning-Enabled {xApps} for Experimental Closed-Loop Optimization in {O-RAN} with {OSC RIC} and {srsRAN}},
  booktitle = {Proc. IEEE ICC Workshops},
  year    = {2025},
}

@article{tsampazi2024pandora,
  author  = {Tsampazi, Maria and D'Oro, Salvatore and Polese, Michele and Bonati, Leonardo and Poitau, Gwenael and Healy, Michael and Alavirad, Mohammad and Melodia, Tommaso},
  title   = {{PandORA}: Automated Design and Comprehensive Evaluation of Deep Reinforcement Learning Agents for Open {RAN}},
  journal = {IEEE Transactions on Mobile Computing},
  year    = {2024},
  doi     = {10.1109/TMC.2024.3505781}
}

@inproceedings{zhao2025adaslicing,
  author    = {Zhao, Ming and Zhang, Yuru and Liu, Qiang and Kak, Ahan and Choi, Nakjung},
  title     = {{AdaSlicing}: Adaptive Online Network Slicing Under Continual Network Dynamics in Open Radio Access Networks},
  booktitle = {Proc. IEEE INFOCOM 2025},
  year      = {2025}
}

@article{intentaware2024,
  author  = {Nahum, Cleverson V. and Lopes, Victor Hugo and Dreifuerst, Ryan M. and Batista, Pedro and Correa, Ilan and Cardoso, Kleber V. and Klautau, Aldebaro and Heath, Jr., Robert W.},
  title   = {Intent-Aware Radio Resource Scheduling in a RAN Slicing Scenario Using Reinforcement Learning},
  journal = {IEEE Transactions on Wireless Communications},
  year    = {2024}
}

@misc{srsran,
  author       = {{Software Radio Systems}},
  title        = {{srsRAN} Project: Open-Source {5G} {CU/DU}},
  howpublished = {\url{https://www.srsran.com/5g}},
  year         = {2025}
}

@inproceedings{foukas2017orion,
  author    = {Foukas, Xenofon and Marina, Mahesh K. and Kontovasilis, Konstantinos},
  title     = {Orion: {RAN} Slicing for a Flexible and Cost-Effective Multi-Service Mobile Network Architecture},
  booktitle = {Proc. ACM MobiCom},
  year      = {2017}
}

@inproceedings{chen2023flexslice,
  author    = {Chen, Chieh-Chun and Chang, Chia-Yu and Nikaein, Navid},
  title     = {{FlexSlice}: Flexible and Real-Time Programmable {RAN} Slicing Framework},
  booktitle = {IEEE Global Communications Conference (GLOBECOM)},
  year      = {2023}
}

@techreport{3gpp_23501,
  author = {{3GPP}},
  title = {System architecture for the {5G} System ({5GS})},
  institution = {3rd Generation Partnership Project},
  type = {Technical Specification},
  number = {TS 23.501},
  note = {Release 17}
}

@inproceedings{wu2025llmxapp,
  title={Llm-xapp: A large language model empowered radio resource management xapp for {5G} {O-RAN}},
  author={Wu, Xingqi and Farooq, Junaid and Wang, Yuhui and Chen, Juntao},
  year={2025},
  publisher={Internet Society}
}

@article{moore2025integrated,
  title={Integrated LLM-Based Intrusion Detection with Secure Slicing xApp for Securing O-RAN-Enabled Wireless Network Deployments},
  author={Moore, Joshua and Abdalla, Aly Sabri and Khanal, Prabesh and Marojevic, Vuk},
  journal={2025 IEEE International Conference on Communications Workshops (ICC Workshops)},
  year={2025},
}

@inproceedings{ai5gtest,
  author  = {Ganiyu, Abiodun and Gajjar, Pranshav and Shah, Vijay K.},
  title   = {{AI5GTest}: {AI}-Driven Specification-Aware Automated Testing and Validation of {5G O-RAN} Components},
  booktitle = {Proc. ACM Conference on Security and Privacy in Wireless and Mobile Networks},
  year    = {2025}
}

@article{oransight,
  author  = {Gajjar, Pranshav and Shah, Vijay K.},
  title   = {{ORANSight-2.0}: Foundational {LLMs} for {O-RAN}},
  journal = {IEEE Transactions on Machine Learning in Communications and Networking},
  year    = {2025}
}

@article{elkael2026allstar,
  title={ALLSTaR—Automated {LLM}-Driven Scheduler Generation and Testing for Intent-Based {RAN}},
  author={Elkael, Maxime and Polese, Michele and Prasad, Reshma and Maxenti, Stefano and Melodia, Tommaso},
  journal={IEEE Transactions on Mobile Computing},
  pages={1-18},
  year={2026}
}

@misc{nvidia_daqiri,
  author       = {{NVIDIA}},
  title        = {{DAQIRI: Data Acquisition for Integrated Real-time Instruments}},
  howpublished = {\url{https://github.com/NVIDIA/daqiri}},
  note         = {Accessed: Aug. 24, 2026}
}

@misc{open5gs,
  author       = {{Open5GS}},
  title        = {{Open5GS}: Open-Source Implementation of {5G} Core and {EPC}},
  howpublished = {\url{https://open5gs.org/}},
  year         = {2026}
}

@misc{ultralytics_yolo,
  author       = {Jocher, Glenn and Qiu, Jing},
  title        = {{Ultralytics} {YOLO11}},
  year         = {2024}
}

@techreport{itu_m2083,
  author = {{ITU-R}},
  title = {{IMT} Vision -- Framework and overall objectives of the future development of {IMT} for 2020 and beyond},
  institution = {International Telecommunication Union},
  type = {Recommendation},
  number = {ITU-R M.2083-0},
  year = {2015},
  month = sep
}

@article{tian2026yolov12,
  title={{YOLOv12}: Attention-centric real-time object detectors},
  author={Tian, Yunjie and Ye, Qixiang and Doermann, David},
  journal={Advances in neural information processing systems},
  volume={38},
  pages={78433--78457},
  year={2026}
}

@inproceedings{hilburn2018sigmf,
  title={SigMF: The signal metadata format},
  author={Hilburn, Ben and West, Nathan and O'Shea, Tim and Roy, Tamoghna},
  booktitle={Proc. GNU Radio Conference},
  volume={3},
  number={1},
  year={2018}
}
\endgroup
\end{document}